\documentclass[preprint,12pt]{elsarticle}

\usepackage{amssymb}
\usepackage{amsthm}
\usepackage{amsmath}
\usepackage{color, soul}

\journal{Engineering Applications of Artificial Intelligence}

\begin{document}

\begin{frontmatter}





\title{New Metric Formulas that Include Measurement Errors in Machine Learning for Natural Sciences}


\author[toelt,hslu]{Umberto Michelucci}

\affiliation[toelt]{organization={TOELT LLC, Artificial Intelligence Research Department},
            addressline={Birchlenstr. 25},
            city={D\"ubendorf},
            postcode={8600},
            country={Switzerland}}

\affiliation[hslu]{organization={Computer Science Department, Lucerne University of Applied Sciences and Arts},
            addressline={Werftestrasse 4},
            city={Lucerne},
            postcode={6002},
            country={Switzerland}}

\author[toelt,zhaw]{Francesca Venturini}

\affiliation[zhaw]{organization={ZHAW School of Engineering},
            addressline={Technikumstrasse 9},
            city={Winterthur},
            postcode={8400},
            country={Switzerland}}

\begin{abstract}
The application of machine learning to physics problems is widely found in the scientific literature. Both regression and classification problems are addressed by a large array of techniques that involve learning algorithms. Unfortunately, the measurement errors of the data used to train machine learning models are almost always neglected. This leads to estimations of the performance of the models (and thus their generalisation power) that is too optimistic since it is always assumed that the target variables (what one wants to predict) are correct. In physics, this is a dramatic deficiency as it can lead to the belief that theories or patterns exist where, in reality, they do not.
This paper addresses this deficiency by deriving formulas for commonly used metrics (both for regression and classification problems) that take into account measurement errors of target variables. The new formulas give an estimation of the metrics which is always more pessimistic than what is obtained with the classical ones, not taking into account measurement errors. The formulas given here are of general validity, completely model-independent, and can be applied without limitations. Thus, with statistical confidence, one can analyze the existence of relationships when dealing with measurements with errors of any kind. The formulas have wide applicability outside physics and can be used in all problems where measurement errors are relevant to the conclusions of studies.
\end{abstract}


\begin{highlights}
\item In this paper, formulas for the expectation value of the metrics MSE, MAE and accuracy that keeps into account measurement errors on the labels in supervised learning are derived.
\item Formulas for the variance of the metrics MSE, MAE and accuracy that keeps into account measurement errors on the labels in supervised learning are derived.
\item Formulas are completely general and are not linked in any way to a specific type of model.
\item The formulas are derived both with a purely statistical, and with a brute-force approach.
\end{highlights}


\end{frontmatter}







\section{Introduction}

The main goal of training a supervised machine learning (ML) model is to find a relationship between a set of $M$ inputs $x_i$ (with $i=1,...,M$) and some outputs $y_i$ (with $i=1,...M$). The values of the variable to be predicted are often called labels in the literature. In general, $x_i$ and $y_i$ can be multidimensional; for simplicity, in this paper, the output variable is assumed to be a real number $y_i\in \mathbb{R}$. In any application of ML to a physics problem the output variables $y_i$ will be either the direct result of a measurement or determined through some calculations from multiple ones. For example, $y_i$ could be the oxygen concentration or temperature of a gas \cite{michelucci2019multi}, oxide glass-forming ability \cite{wilkinson2022hybrid}, temperature in melt-pool fluid dynamics \cite{zhu2021machine}, or a measure of the dissolution kinetics of gases \cite{krishnan2018predicting}.

ML is used in physics in a large number of cases. For example, to locate phase transitions without any physical knowledge \cite{carrasquilla2017machine, morningstar2017deep, tanaka2017detection}, to select events in collisions \cite{baldi2016jet, de2016jet} and to flavour tagging \cite{guest2016jet} in particle physics. In cosmology, ML has been applied, for example, to estimate the photometric redshift \cite{carrasco2013tpz, collister2007megaz} and to predict fundamental cosmological parameters based on the dark matter spatial distribution \cite{ravanbakhsh2016estimating}. The list of examples is incredibly long, and applications can be found in almost all fields of physics. For an extensive review, the interested reader is referred to \cite{carleo2019machine}.

In all cases, to the best of the authors' knowledge, the performance of the models is reported without taking into account the measurement errors on the labels. Target variables that have errors present a certain uncertainty, and it is not clear how this uncertainty propagates to the metrics used to measure the performance of ML models.
Research starts to indicate that in many physics problems, ignoring measurement errors may lead to an underestimation of ML model uncertainties \cite{ghosh2022cautionary}.

The problem is of fundamental relevance since typically models are trained on a specific dataset, typically split into a training and test part, and results are given in terms of specific metrics, such as the mean squared error (MSE), the mean absolute error (MAE) for regression problems or the accuracy ($a$) for classification problems (or slightly different metrics in case of unbalanced datasets of multi-class classification problems). The problem is that in almost all cases, measurement errors on the variables to be predicted ($y_i$) are ignored. This will lead to overly optimistic estimates of the mentioned metrics (such as the MSE, MAE or $a$).

 The problem of noisy labels (a different kind of problem than the one analyzed in this paper) is a relatively widely researched topic \cite{pmlr-v37-menon15, cour2011learning, pmlr-v119-zheng20c, natarajan2013learning, pmlr-v119-yao20b}. Some articles deal with methods to identify wrongly labelled observation \cite{pmlr-v119-bahri20a} and some try to define modified loss functions that can deal with noise \cite{7159100}. All efforts are directed towards understanding how to get better model performance or model stability when labels are noisy. However, none of these works addresses the problem of how labels errors can influence the metrics evaluated. Particularly in physics (but in all sciences for that matter), measurement errors must be included in any results any physicist learns early in his or her career. This paper addresses this deficiency and provides formulas that take into account errors to better estimate the metrics most commonly used in ML, both for regression and classification problems.

The main contributions of this paper are four.
Firstly, formulas are given for the MSE, the MAE, the accuracy $a$, and their variances that take into account measurement errors on the variable to be predicted $y_i$. These formulas are of general validity and are independent of the ML model used.
Secondly, all the formulas are fully derived mathematically with a statistical approach. Thirdly, an a-priori mathematical derivation for the formulas is given in the appendices of the paper. Finally, guidelines are presented on how to use those formulas.

\section{Problem Formulation and Notation}

This paper considers the following thought experiment: find a relationship between a set of $M$ inputs $x_i$ (with $i=1,...,M$) and some outputs $y_i$ (with $i=1,...M$) that are assumed to be independent.
The typical steps to achieve this are to split the dataset into training and test datasets, train the model on the training dataset, and validate the results by applying the model to the test dataset, namely, on unseen data.
The training is done by minimizing an appropriate loss function, typically the MSE or MAE for regression, or the cross entropy for classification. Several metrics can be evaluated to asses the performance of the trained model. In this paper, the metrics most commonly used are discussed: MSE, MAE and accuracy $a$.

An important step to evaluate the generalisation properties of the trained model is to perform a cross-validation \cite{michelucci2021estimating}.
An example of cross-validation is the split-train approach, achieved by performing the dataset split multiple times thus obtained multiple training and test datasets. Each time a new model is trained and its performance evaluated on the test dataset. Naturally, every time a new split is used, the model changes. Looking at the metrics obtained by evaluating them on the multiple test datasets, one can get an indication on the average performance of possible models obtained with the initial datasets.
Unfortunately, this method does not take into account in any way the errors that are inevitably present on the $y_i$. Therefore, the metrics' values will lead to an overly optimistic impression of the goodness of the model.

Let us introduce a measurement error on the labels $y_i$ with the random variable $\epsilon_i$
\begin{equation}
    y_i = \overline{y}_i +\epsilon_i
\end{equation}
where $\overline{y}_i$ is the average value of $y_i$ and $\epsilon_i$ follows a normal distribution with average zero and variance $\sigma_i^2$, or more formally,
\begin{equation}
    \epsilon_i \sim \mathcal{N}(0, \sigma_i^2)
\end{equation}
or in other words
\begin{equation}
    y_i \sim \mathcal{N}(\overline{y}_i, \sigma_i^2).
    \label{eq:nordist}
\end{equation}
This is based on the hypothesis that measurement errors follow a Gaussian distribution \cite{taylor1997introduction}.
This work provides formulas for the expected value and variance of metrics, specifically MSE, MAE, and $a$, over the distribution of the random variable $\epsilon_i$.

In this paper, two generic problems are considered: a regression problem and a classification problem.
A generic \textbf{regression problem} has the objective of predicting a continuous variable using a set of $M$ observations tuples $(x_i, y_i)$, with $i=1,...M$. $x_i$ is the $i^{th}$ input, and $y_i$ is the $i^{th}$ target variable. In this case, the two metrics considered here are the mean squared error (MSE)
\begin{equation}
\textrm{MSE} = \frac{1}{M}\sum_{i=1}^M (y_i-\hat y_i)^2
\label{MSE_no_error}
\end{equation}
and the mean absolute error (MAE)
\begin{equation}
\textrm{MAE} = \frac{1}{M}\sum_{i=1}^M |y_i-\hat y_i|
\label{MAE_no_error}
\end{equation}
where $\hat y_i$ indicates the prediction of the ML model.

A generic \textbf{binary classification problem} has the objective of classifying a set of $M$ observations $x_i$ with $i=1,...,M$ into two classes $1$ and $0$. The class labels to be predicted are indicated, as in the regression problem, with $y_i\in \{0,1\}$. In this case, the most used metric is the accuracy ($a$) obtained simply by
\begin{equation}
    a = \frac{\textrm{Number of correctly classified observations}}{M}
    \label{a_no_error}
\end{equation}
The impact of unbalanced datasets on the accuracy is not discussed here and the reader is referred to other works \cite{michelucci2018applied}.

The estimation of the expected value and variance of the metrics of Equation (\ref{MSE_no_error})-(\ref{a_no_error}) that account for the uncertainty $\epsilon_i$ on $y_i$ will be given in Sect. \ref{sec:regression} and \ref{sec:classification} for the regression and classification problems respectively.

\section{Regression Problem}
\label{sec:regression}

\subsection{Mean Squared Error (MSE) Estimate}

As discussed earlier, the target variables measurements $y_i$ are assumed independent and following a normal distribution (this is a common assumption when dealing with measurement errors) with average ${\overline y}_i$ and standard deviation $\sigma_i$ (see Equation \ref{eq:nordist}).
The standard deviation $\sigma_i$ may differ for different $i$, for example if $y_i$ has different errors in different ranges of its value.
The standard deviation $\sigma_i$ is a way of giving an estimate of the measurement error.

Let's consider first he special case when $\sigma_i = \sigma$ for $i=1,...,M$ (in other words, when the standard deviation of the $y_i$ is constant). In this case, there is an elegant way to understand everything about the MSE behaviour without any complex calculations. This is reported in Subsection \ref{sec:stat}.
The general formulas that apply for $\sigma_i$ not constant are given in Subsection \ref{sec:direct}.

\subsubsection{Constant variances $\sigma_i^2$}
\label{sec:stat}

This section covers the case where $\sigma_i = \sigma$ for $i=1,...,M$. Let us rewrite the MSE as
\begin{equation}
\label{eq:deltasigma}
    \textrm{MSE} = \frac{\sigma^2}{M} \sum_{i=1}^M \left(
        \frac{\delta_i}{\sigma}
    \right)^2
\end{equation}
where $\delta_i \equiv y_i-\hat y_i$. Note that since $\delta_i$ are normally distributed, clearly
\begin{equation}
    \frac{\delta_i}{\sigma} \sim {\mathcal N}\left(
    \frac{{\overline \delta}_i }{\sigma}, 1 \right)
\end{equation}
Equation (\ref{eq:deltasigma}) is the sum of normally distributed random variables squared with a variance of one, and, therefore, their sum follows the non-central chi-squared distribution with $M$ degrees of freedom (indicated here with $\chi'_M(\lambda)$) \cite{hogg2010probability}
\begin{equation}
\label{eq:lambda1}
    \sum_{i=1}^M \left(
        \frac{\delta_i}{\sigma} \right)^2 \sim \chi'_M(\lambda)
\end{equation}
with the noncentrality parameter $\lambda$ given by
\begin{equation}
    \lambda =  \sum_{i=1}^M \frac{\overline \delta_i^2}{\sigma^2}
\end{equation}
where $\overline \delta_i = \overline y_i - \hat y_i$
Thus from Equation (\ref{eq:deltasigma}) and (\ref{eq:lambda1}) we can say that the MSE satisfies
\begin{equation}
\label{eq:mse1}
    \frac{M}{\sigma^2} \textrm{MSE} \sim \chi'_M(\lambda).
\end{equation}
After knowing this, it is straightforward to evaluate
\begin{equation}
    \mathbb{E}\left(\frac{M}{\sigma^2}\textrm{MSE} \right)
\end{equation}
and
\begin{equation}
    \textrm{Var}\left(\frac{M}{\sigma^2}\textrm{MSE}\right)
\end{equation}
In fact, it is a well-known result that for a random variable $X$ that follows a noncentral chi-square distribution $\chi'_M(\lambda)$ \cite{hogg2010probability} it is true that
\begin{equation}
\label{eq:var2}
    \mathbb{E}(X) = M+\lambda
\end{equation}
and
\begin{equation}
\label{eq:variance1}
    \textrm{Var}(X) = 2M+4\lambda.
\end{equation}
With the help of Equations (\ref{eq:mse1}) and (\ref{eq:variance1}) the expected value of the MSE can be rewritten in a compact and quite interpreteble form:
\begin{equation}
\label{eq:sigmaeq}
    \mathbb{E}\left(\frac{M}{\sigma^2}\textrm{MSE}
    \right)=M+\lambda
\end{equation}
and, therefore,
\begin{equation}
\mathbb{E}(\textrm{MSE})  = \frac{1}{M} \sum_{i=1}^M({\overline y}_i - \hat y_i)^2+\sigma^2.
    \label{Eq:E_MSE_const_var}
\end{equation}
Equation (\ref{Eq:E_MSE_const_var}) indicates that a better estimation of $\mathbb{E}$ is obtained by the MSE evaluated with the average of the labels plus variance of the measurements $y_i$.

The formula for the variance of the MSE is given, using Equation (\ref{eq:variance1}), by the formula
\begin{equation}
\label{eq:var1}
    \textrm{Var}(\textrm{MSE}) = \frac{2\sigma^4}{M}+\frac{4\sigma^2}{M^2}\sum_{i=1}^M
    \overline \delta_i^2.
\end{equation}

Note that the formulas given are only valid in the case where the variances of the single $y_i$ are equal to a constant $\sigma$. This is not always the case, and especially in real life cases, quantities may have different measurement errors depending on their values. The general case is discussed in the next section.

\subsubsection{Non-constant variances $\sigma_i^2$}
\label{sec:direct}

To evaluate the MSE expected value and its variance in the case where the $\sigma_i$ are all different, one can use two approaches: a statistical one and an a-priori one that consist in  evaluating the necessary integrals directly. The statistical approach is described in this section. The a priori in \ref{app:reg_MSE}.
For non-constant variances, it is impossible to reduce the sum
\begin{equation}
    \frac{1}{M}\sum_{i=1}^M \delta_i^2
\end{equation}
to a sum of $M$ variables with different averages but unit variances (the prerequisites to get the noncentral chi-square distribution used in the previous section). In fact, in this case, in general
\begin{equation}
    \delta_i \sim {\mathcal N}\left(
    {\overline \delta}_i , \sigma_i^2 \right)
\end{equation}
and, therefore, we cannot use the same strategy that was used in the previous section.
To determine the expected value and variance, let's observe that
\begin{equation}
    \frac{\delta_i}{\sigma_i} \sim {\mathcal N}\left(
    \frac{{\overline \delta}_i}{\sigma_i} , 1 \right)
\end{equation}
thus
\begin{equation}
    \left(\frac{\delta_i}{\sigma_i} \right)^2\sim
    \chi'_1(\lambda)
\end{equation}
with $\lambda = {\overline \delta}_i^2/\sigma_i^2$ is the noncentrality parameter. Using the expected value formula for a non-central chi-squared distribution with one degree of freedom the expectation value is
\begin{equation}
\label{eq:exp1}
  \mathbb{E}\left[\left(\frac{\delta_i}{\sigma_i} \right)^2
    \right] =1+\frac{{\overline \delta}_i^2}{\sigma_i^2} \ \ \Rightarrow \ \
    \mathbb{E}(\delta_i^2) =\sigma_i^2+{\overline \delta}_i^2
\end{equation}
Now
\begin{equation}
\label{eq:mse2}
   \mathbb{E}(\textrm{MSE}) =  \mathbb{E}\left(\frac{1}{M}\sum_{i=1}^M \delta_i^2\right)
\end{equation}
and substituting Equation (\ref{eq:exp1}) in Equation (\ref{eq:mse2})
\begin{equation}
\begin{aligned}
    \mathbb{E}(\textrm{MSE}) &= \mathbb{E}\left(\frac{1}{M}\sum_{i=1}^M \delta_i^2\right)
    =\frac{1}{M} \sum_{i=1}^M \mathbb{E}(\delta_i^2)= \\
    &=\frac{1}{M} \sum_{i=1}^M ({\overline \delta}_i^2 + \sigma_i^2)
\end{aligned}
\end{equation}
that is the generalized version of Equation (\ref{Eq:E_MSE_const_var}):
\begin{equation}
    \mathbb{E}(\textrm{MSE}) =\frac{1}{M}
    \sum_{i=1}^M ({\overline \delta}_i^2 + \sigma_i^2).
\end{equation}
Let us turn our attention to the variance. From Equation (\ref{eq:variance1} with $M=1$ it follows that
\begin{equation}
    \textrm{Var}\left(
        \frac{\delta_i^2}{\sigma_i^2}
    \right) = 2 + 4\frac{{\overline \delta}_i^2}{\sigma_i^2}
\end{equation}
and by using the property that
\begin{equation}
    \textrm{Var}\left(
        \frac{\delta_i^2}{\sigma_i^2}
    \right) = \frac{1}{\sigma_i^4} \textrm{Var}(\delta_i^2)
\end{equation}
the variance becomes
\begin{equation}
\label{eq:vardelta1}
    \textrm{Var}(\delta_i^2) = 2\sigma_i^4+4 {\overline \delta}_i^2 \sigma_i^2
\end{equation}
and thus since
\begin{equation}
    \textrm{Var}(\textrm{MSE}) = \textrm{Var}\left(\frac{1}{M} \sum_{i=1}^M  \delta_i^2\right)
    = \frac{1}{M^2} \sum_{i=1}^M
    \textrm{Var}(\delta_i^2)
\end{equation}
 using Equation (\ref{eq:vardelta1})
\begin{equation}
    \textrm{Var}(\textrm{MSE})=\frac{2}{M^2} \sum_{i=1}^M
    \sigma_i^4 +\frac{4}{M^2} \sum_{i=1}^M {\overline \delta}_i^2 \sigma_i^2
\end{equation}
that is the generalized version of Equation (\ref{eq:var1}).

The a priori determination of $\mathbb{E}(\textrm{MSE})$ is performed by evaluating the integral
\begin{equation}
    \mathbb{E}(\textrm{MSE}) = \frac{1}{M} \sum_{i=1}^M\left[
    \frac{1}{\sqrt{2\pi}\sigma_i} \int_\mathbb{R} (y_i - \hat y_i)^2 e^{-\frac{(y_i - \overline y_i)^2}{2\sigma_i^2}} dy_i
    \right]
\end{equation}
This calculation requires some work and is shown in \ref{app:reg_MSE} for completeness.


\subsection{Mean Absolute Error (MAE) Estimate}

Let us turn our attention to the MAE. Since $y_i \sim \mathcal{N}({\overline{y}_i, \sigma_i^2})$ the quantity
\begin{equation}
\label{eq:gammai}
   |\delta_i| = |y_i - \hat y_i|
\end{equation}
follows a folded normal distribution, indicated here with $\mathcal{F}({\overline{\delta}_i, \sigma_i^2})$. The expected value of a random variable $X$ following a folded distribution, $X\sim \mathcal{F}({\mu, \sigma^2})$ is given by \cite{hogg2010probability}
\begin{equation}
    \mathbb{E}(X) = \mu\sqrt{\frac{2}{\pi}}e^{-\mu^2/(2\sigma^2)}
    +\mu\left(1-2\Phi\left(-\frac{\mu}{\sigma}
    \right)\right)
\end{equation}
where $\Phi$ is the normal cumulative distribution function. The same formula can be expressed in terms of the error function as
\begin{equation}
\label{eq:ex1}
    \mathbb{E}(X) = \mu\sqrt{\frac{2}{\pi}}e^{-\mu^2/(2\sigma^2)}
    +\mu\ \textrm{erf}\left(\frac{\mu}{\sqrt{2}\sigma} \right).
\end{equation}

The expected value of MAE is given by
\begin{equation}
    \mathbb{E}(\textrm{MAE}) = \mathbb{E}\left(\frac{1}{M}\sum_{i=1}^M |\delta_i|\right) =\frac{1}{M}\sum_{i=1}^M\mathbb{E}(|\delta_i|)
\end{equation}
the expected value is given by
\begin{equation}
\begin{array}{lll}
     \mathbb{E}(\textrm{MAE}) & = &  \displaystyle\frac{1}{M}\displaystyle\sum_{i=1}^M |\hat y_i-\overline{y}_i|+\frac{1}{M}\sum_{i=1}^M \left\{
      \frac{\sqrt{2}}{\sqrt{\pi}} \sigma_i e^{-\delta_i^2/(2\sigma_i^2)} +\right.\\
     & & -\left. |\delta_i| \textrm{erfc}\left(
        \displaystyle \frac{|\delta_i|}{\sqrt{2}\delta_i}
     \right)\right\}\\
\end{array}
\end{equation}
where Equation (\ref{eq:ex1}) has been rewritten using the function $\textrm{erfc}()$ to write $\mathbb{E}(\textrm{MAE})$ as the sum of the known formula for $MAE$ (albeit evaluated with the averages of $y_i$) plus a correction term $\Delta(MAE)$
\begin{equation}
    \mathbb{E}(\textrm{MAE})  =   \displaystyle\frac{1}{M}\displaystyle\sum_{i=1}^M |\hat y_i-\overline{y}_i|+\Delta(MAE)
\end{equation}
with
\begin{equation}
    \Delta(\textrm{MAE}) = \frac{1}{M}\sum_{i=1}^M \left\{
      \frac{\sqrt{2}}{\sqrt{\pi}} \sigma_i e^{-\delta_i^2/(2\sigma_i^2)} -
       |\delta_i| \textrm{erfc}\left(
        \displaystyle \frac{|\delta_i|}{\sqrt{2}\delta_i}
     \right)\right\}
\end{equation}
Finally, the variance of the MAE $\textrm{Var}(\textrm{MAE})$ is analyzed.
The variance of a random variable $X$ such that $X\sim \mathcal{F}({\mu, \sigma^2})$ is given by
\begin{equation}
    \textrm{Var}(X)=\mu^2+\sigma^2-\overline{X}^2
\end{equation}
Considering $X=|\delta_i|$
\begin{equation}
\label{eq:vargammai}
    \textrm{Var}(|\delta_i|)={\overline \delta}_i^2+\sigma_i^2-\overline{\delta}_i\sqrt{\frac{2}{\pi}}e^{-\overline{\delta}_i^2/(2\sigma^2)}
    -\overline{\delta}_i\textrm{erf}\left(\frac{\overline{\delta}_i}{\sqrt{2}\sigma} \right)
\end{equation}
Since the different measurements $i$ are independent, the property $\textrm{Var}(X+Y)=\textrm{Var}(X)+\textrm{Var}(Y)$ can be used (in fact, in this case $\textrm{Cov}(X,Y)=0$). Thus, by using Equation (\ref{eq:vargammai}) one can derive the following formula
\begin{equation}
\label{eq:varmae}
    \textrm{Var}(\textrm{MAE}) = \frac{1}{M^2} \sum_{i=0}^M \left\{ {\overline \delta}_i^2+\sigma_i^2-\overline{\delta}_i\sqrt{\frac{2}{\pi}}e^{-\overline{\delta}_i^2/(2\sigma^2)}
    -\overline{\delta}_i\textrm{erf}\left(\frac{\overline{\delta}_i}{\sqrt{2}\sigma} \right)
    \right\}
\end{equation}

\section{Classification Problem}
\label{sec:classification}

In a binary classification problem the machine learning model typically outputs the probability $\hat y_i$ of an observation of being in class 1 (the labels are $y_i \in \{ 0,1 \}$). The conversion of the probability into a class is then done using the Heaviside step function $H(x-\alpha)$ where $\alpha$ is a treshold that is normally chosen as $\alpha = 1/2$. The accuracy $a$ can be written for a binary problem in the following form
\begin{equation}
\label{eq:acc_cp}
    a = \frac{1}{M} \sum_{i=1}^M \left[
        y_i H(\hat y_i-\alpha)+(1-y_i)(1- H(\hat y_i-\alpha))
    \right]
\end{equation}
The two terms appearing in Equation (\ref{eq:acc_cp}) correspond to the true positives (TP) and true negatives (TN). In facts
\begin{equation}
    \begin{array}{lll}
        \textrm{TP} & = & \displaystyle \frac{1}{M} \sum_{i=1}^M \left[
        y_i H(\hat y_i-\alpha) \right]\\
        \textrm{TN} & = & \displaystyle \frac{1}{M} \sum_{i=1}^M \left[
        (1-y_i)(1- H(\hat y_i-\alpha) \right].
    \end{array}
\end{equation}
Let us consider the case where there is a probability $q<1$ that an observation label $y_i$ is wrong.
Let us start by considering $b_j$, a Bernoulli random variable with a probability $p$ of being one (and consequently a probability $q=1-p$ of being 0). Let us define the random variable
\begin{equation}
    _r y_i=y_i b_j+(1-b_j)(1-y_i).
\end{equation}
$_r y_i$ will assume the value $y_i$ with a probability $p$. $_r y_i$ will assume the value of $1 - y_i$  with a probability $q$. The expectation value and the variance of $_r y_i$ are given by
\begin{equation}
\label{eq:expy}
\begin{aligned}
    \mathbb{E}(_r y_i) =& y_i \mathbb{E}(b_j)+(1-y_i)\mathbb{E}(1-b_j)=y_ip+(1-y_i)q=\\
    =&y_i(1-2q)+q
\end{aligned}
\end{equation}
and
\begin{equation}
\label{eq:vary}
\begin{aligned}
    \textrm{Var}(_r y_i) =& y_i^2 \textrm{Var}(b_j)+(1-y_i)^2\textrm{Var}(1-b_j) = \\
    =&y_1pq+(1-y_i)pq=pq
\end{aligned}
\end{equation}
Equation (\ref{eq:vary}) can be derived  by noting that since $y_i\in\{0,1\}$ it is true that $y_i^2=y_i$ and $(1-y_i)^2=(1-y_i)$.
The accuracy in the presence of errors in the labels is obtained by using the random variable $_r y_i$ in Equation (\ref{eq:acc_cp}), which results in
\begin{equation}
\label{eq:acc}
    _r a = \frac{1}{M} \sum_{i=1}^M \left[
        _r y_i H(\hat y_i-\alpha)+(1-_r y_i)(1- H(\hat y_i-\alpha))
    \right]
\end{equation}
Now all ingredients are available to calculate $\mathbb{E}(_r a)$ and $\textrm{Var}(_r a)$ with the help of Equations (\ref{eq:expy}) and (\ref{eq:vary}).
Using the properties of the expected value and of the variance, the following results can be obtained in just a few steps
\begin{equation}
\label{eq:finalacc}
\begin{cases}
    \mathbb{E}(_r a) = a + q(1-2a) \\
    \textrm{Var}(_r a) = pq = (1-q)q.
\end{cases}
\end{equation}

In \ref{app:class} an alternative and more intuitive way of obtaining $\mathbb{E}(_r a)$ is described.

Equation \ref{eq:finalacc} is a very interesting result that needs some discussion. First of all, it can be observed that for any model for which $a>1/2$, $\mathbb{E}(_ra)<a$, as expected. The errors of the labels effectively reduce the performance of the model. Note that any model in a binary classification problem that has $a<1/2$ can be transformed into one with $a>1/2$ by simply exchanging all predictions: 1 into 0 and vice versa.

The expected value of the accuracy, can also be written in a more compact form as
\begin{equation}
    \mathbb{E}(_r a) = \displaystyle \frac{1}{M} \sum_{i=1}^M \left[
        (1-q) \mathcal{A}(y_i)+ q \mathcal{A}(1-y_i) \right]
\end{equation}
where $\mathcal{A}(y_i)$ is
\begin{equation}
    \mathcal{A}(y_i)=y_i H(\hat y_i-\alpha)+(1-y_i)(1- H(\hat y_i-\alpha))
\end{equation}
from Equation (\ref{eq:acc}).
To better understand this formula, one needs to rewrite it a slightly different form by using Equation (\ref{eq:app1}) derived in \ref{app:reg_MAE}, reported here for clarity
\begin{equation}
    \mathbb{E}(_r a) = (1-q) \frac{\textrm{TP}+\textrm{TN}}{M} + q \frac{\textrm{FP}+\textrm{FN}}{M}
\end{equation}
This formula can be interpreted with the help of Figure \ref{fig:figure1}:
\begin{figure}
    \centering
    \includegraphics[width=14cm]{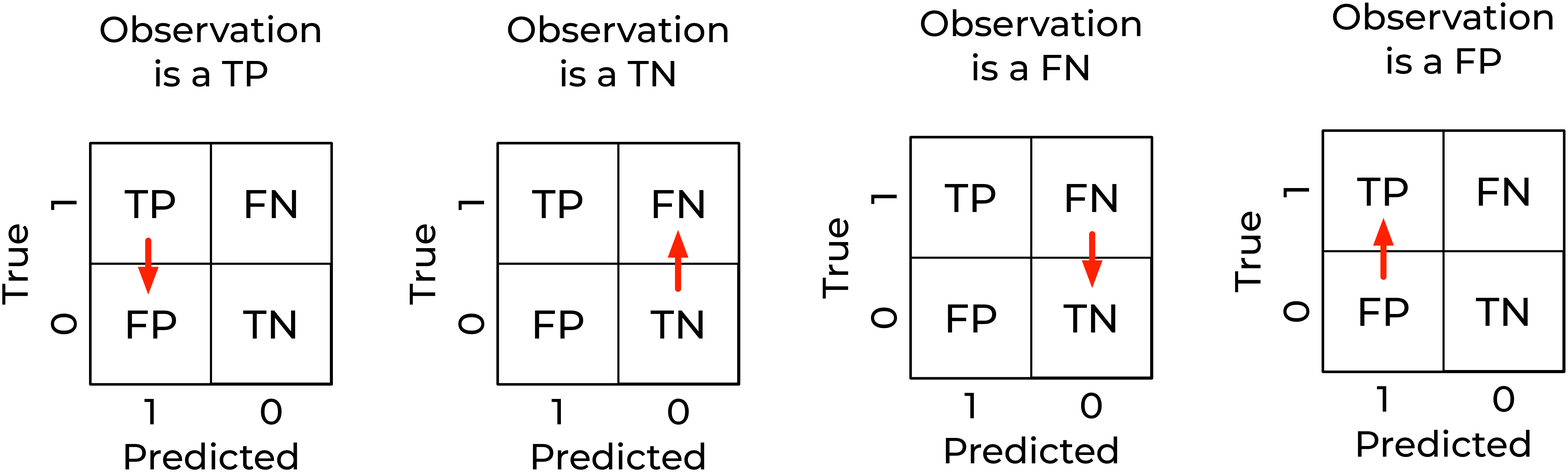}
    \caption{A visual representation of the effect of a wrong label for an observation: the red arrow indicates how the observation will be classified in the confusion matrix. TP: true positives, TN: true negatives, FP: false positives and FN: false negatives.}
    \label{fig:figure1}
\end{figure}
\begin{itemize}
    \item If an observation is a true positive or a true negative and the label is wrong, it will be classified with a probability $q$ as either a false positive or false negative, respectively. In other words, the error of the label will reduce the number on the diagonal of the confusion matrix and will contribute to the off-diagonal terms.
    \item Analogously, if an observation is a false negative or false positive and the label is wrong, the error on the label has the effect of reducing the off-diagonal term of the confusion matrix and contributes to the diagonal terms.
\end{itemize}
Equation (\ref{eq:finalacc}) gives an estimate of accuracy taking into account a possible error in the labels, due for example to measurement errors. The simple accuracy $a$ obtained neglecting measurement errors is an overestimation and does not give a correct picture of how a model could perform.

Let us consider for example a hypothetical model that has obtained $a=0.85$ with labels with a probability of 5\% of being wrong. Equation \ref{eq:finalacc} gives an estimate of the accuracy of $\mathbb{E} (_r a) =\textrm{0.815\%}$. The difference is not negligible and must be taken into account for any application of machine learning to scientific results that involves measurement errors (basically always).

\section{How to use the Formulas}

Tables \ref{tab:exp2} and \ref{tab:var2} summarize the formulas derived in this paper. Note that the application of formulas requires not only the standard deviation of the measurements $\sigma_i$, but also its average ${\overline y}_i$.
When measuring a quantity $y_i$ the scientist should try to get enough measurements to be able to estimate average and standard deviation. Measurements may be expensive, and therefore could lead to having only a limited number of them, making estimating average and standard deviation difficult or, in extreme cases, impossible with a statistical approach. In such a case, the standard deviation should be replaced with the measurement error obtained by a classical propagation of the experimental errors. The average should be evaluated by calculating the mean of the few available values of $y_i$, or, in the extreme case where only one value is available, using this value in place of the average.

When calculating the MSE, MAE or accuracy the following process should be followed:
\begin{enumerate}
    \item Multiple measurements for the $y_i$ should be performed to be able to evaluate $\overline{y_i}$. $\sigma_i$ can be estimated statistically (by evaluating the variance of the multiple measurements of $y_i$) or by doing error propagation by using the knowledge about how the measurement were performed. The latter way is more practical. In fact to get a good estimate of errors by statistical means one needs a large amount of measurements, and that is not always possible.
    \item The appropriate metric from Table \ref{tab:exp2} (depending on the kind of problem one is trying to solve) should be chosen and calculated according to the formula.
    \item If needed, the variance corresponding to the chosen metric should be chosen from Table \ref{tab:var2}.
\end{enumerate}
The use of the formulas described in this paper will give a more realistic estimate of machine learning metrics (here MSE, MAE and accuracy) and therefore of the model performance since it takes into account measurement errors on the target variables.

\begin{table}[hbt]
\centering
\begin{tabular}[t]{l l}
 \hline
 \\[-0.4cm]
 \textbf{Metric} & \textbf{Formula} \\ [0.5ex]
 \hline
 MSE &
 \begin{tabular}{l}$\mathbb{E}(\textrm{MSE})  = \displaystyle \frac{1}{M} \sum_{i=1}^M({\overline y}_i - \hat y_i)^2+\sigma^2$  \\ Case when $\sigma_i = \sigma$ for $i=1,...,M$\end{tabular} \\[1.2cm]
 MSE & \begin{tabular}{l}$\mathbb{E}(\textrm{MSE}) =\displaystyle\frac{1}{M}
    \sum_{i=1}^M (({\overline y}_i - \hat y_i)^2 + \sigma_i^2)$  \\
    For $\sigma_i$ not constant\\
\end{tabular} \\[1.2cm]
 MAE &   \begin{tabular}{l}$\mathbb{E}(\textrm{MAE})= \displaystyle\frac{1}{M}\displaystyle\sum_{i=1}^M |\hat y_i-\overline{y}_i|$+\\
 $\displaystyle \frac{1}{M}\sum_{i=1}^M \left\{
      \frac{\sqrt{2}}{\sqrt{\pi}} \sigma_i e^{-\delta_i^2/(2\sigma_i^2)} -
      |\delta_i| \textrm{erfc}\left(
        \displaystyle \frac{|\delta_i|}{\sqrt{2}\delta_i}
     \right)\right\}$ \\[0.7cm] \end{tabular} \\
     &For $\sigma_i$ not constant\\[0.6cm]
 accuracy & $\mathbb{E}(_r a) = a + q(1-2a)$  \\[1ex]
 \hline
\end{tabular}
\caption{Formulas for the expected value of MSE, MAE and accuracy ($a$) that take into account errors on the target variables.}
\label{tab:exp2}
\end{table}

\begin{table}[hbt]
\centering
\begin{tabular}[t]{ll}
 \hline
 \\[-0.4cm]
 \textbf{Metric} & \textbf{Formula} \\ [0.5ex]
 \hline
 \\[-0.4cm]
 MSE & $\textrm{Var}\left(\textrm{MSE}\right)
    = \displaystyle\frac{2}{M^2} \sum_{i=1}^M
    \sigma_i^4 +\frac{4}{M^2} \sum_{i=1}^M {\overline \delta}_i^2 \sigma_i^2$ \\[1.2cm]
 MAE &
 $\textrm{Var}(\textrm{MAE}) =
 \displaystyle\frac{1}{M^2} \sum_{i=0}^M \left\{ {\overline \delta}_i^2+\sigma_i^2-\overline{\delta}_i\sqrt{\frac{2}{\pi}}e^{-\overline{\delta}_i^2/(2\sigma^2)}
    -\overline{\delta}_i\textrm{erf}\left(\frac{\overline{\delta}_i}{\sqrt{2}\sigma} \right)
    \right\}$
  \\[0.7cm]
  accuracy & $\textrm{Var}(_r a)=pq=q(1-q)$  \\[0.4cm]
 \hline
\end{tabular}
\caption{Formulas for the variance of the MSE, MAE and accuracy ($a$).}
\label{tab:var2}
\end{table}

\section{Conclusions}

This work presents for the first time formulas for calculating the metrics commonly used in ML, namely MSE, MAE, and accuracy, taking into account the errors in the target variables. The formulas, which are of general validity, are derived using both a statistical and an a priori approach. They give more realistic estimates of the metrics that are otherwise overly optimistic.
The analysis shows that the MSE and MAE calculated with the derived formulas are always larger than the one obtained ignoring errors in the measurements (in other words, setting $\sigma_i=0$ for $i=1,...,M$). The accuracy evaluated according to the formula given in Table \ref{tab:exp2} is always lower than the one evaluated by using only the target variables and ignoring possible errors.

Another important contribution of this paper is that it shows the relevance of performing multiple repeated  measurements to calculate averages and variances of measurements. These are crucial to obtain scientifically accurate estimates of ML metrics, and therefore, ML model performances.
The reported formulas have a very wide applicability and should be used any time the target variables are known within an error an error.

\appendix

\section{Direct Calculation of $\mathbb{E}(\textrm{MSE})$}
\label{app:reg_MSE}

The integral to be evaluated is
\begin{equation}
\label{eq:reg_integral}
    \mathbb{E}(\textrm{MSE}) = \frac{1}{M} \sum_{i=1}^M\left[
    \frac{1}{\sqrt{2\pi}\sigma_i} \int_\mathbb{R} (y_i - \hat y_i)^2 e^{-\frac{(y_i - \overline y_i)^2}{2\sigma_i^2}} dy_i
    \right]
\end{equation}
To solve this integral, the following change of variable can be used
\begin{equation}
    s = \frac{y_i-\overline y_i}{\sqrt{2}\sigma_i}
\end{equation}
this of course leads to
\begin{equation}
    dy_i = ds\sqrt{2}\sigma_i .
\end{equation}
Therefore, Equation (\ref{eq:reg_integral}) can be rewritten as
\begin{equation}
     \frac{1}{\sqrt{\pi} M} \sum_{i=1}^M\int_\mathbb{R} (s\sigma_i \sqrt{2}+{\overline y}_i-\hat y_i)^2e^{-s^2}ds
\end{equation}
and by expanding the polynomial squared and defining $\delta_i = \hat y_i - {\overline y}_i$ one obtains
\begin{equation}
\label{eq:3terms}
    \frac{1}{\sqrt{\pi} M} \sum_{i=1}^M\int_\mathbb{R} (2s^2\sigma_i^2+\delta_i^2 - 2\sqrt{2}\sigma_i s \delta_i)e^{-s^2}ds
\end{equation}
in Equation (\ref{eq:3terms}) there are three terms that need to be evaluated.
\begin{equation}
    \mathbb{E}(\textrm{MSE}) = A+B+C
\end{equation}
with
\begin{equation}
  \begin{array}{ccc}
     A & = &\displaystyle \frac{2}{\sqrt{\pi}M} \sum_{i=1}^M \sigma_i^2\int_\mathbb{R} s^2 e^{-s^2}ds \\
     B & = & \displaystyle \frac{1}{\sqrt{\pi}M} \sum_{i=1}^M \delta_i^2 \int_\mathbb{R} e^{-s^2}ds\\
     C & = & -\displaystyle \frac{2 \sqrt{2}}{\sqrt{\pi}M} \sum_{i=1}^M \sigma_i \delta_i \int_\mathbb{R}   s e^{-s^2}ds
  \end{array}
\end{equation}
given the symmetry of the function under the integral sign in $C$ it is immediately evident that $C=0$.  Using the results,
\begin{equation}
    \int_\mathbb{R} e^{-s^2}ds = \sqrt{\pi}
\end{equation}
and
\begin{equation}
    \int_\mathbb{R} s^2 e^{-s^2}ds = \frac{\sqrt{\pi}}{2}
\end{equation}
$A$ and $B$ can be easily calculated
\begin{equation}
  \begin{array}{ccc}
     A & = &\displaystyle \frac{1}{M} \sum_{i=1}^M \sigma_i^2 \\
     B & = & \displaystyle \frac{1}{M} \sum_{i=1}^M \delta_i^2
  \end{array}
\end{equation}
Note how $B$ is the  MSE evaluated with the measurement averages ${\overline y}_i$, while $A$ is the average of the measurement standard deviations. So Equation (\ref{eq:3terms}) can be finally rewritten as
\begin{equation}
    \mathbb{E}(\textrm{MSE}) = \displaystyle \frac{1}{M} \sum_{i=1}^M \sigma_i^2 + \displaystyle \frac{1}{M} \sum_{i=1}^M ({\overline y}_i - \hat y_i)^2
\end{equation}
This concludes the derivation. The calculation of $\textrm{Var}(\textrm{MSE})$ is not reported here, as it is similar to the one for the expected value and would make this paper unbearably long.

\section{Direct Calculation of $\mathbb{E}(\textrm{MAE})$}
\label{app:reg_MAE}

The integral to be evaluated is
\begin{equation}
    \mathbb{E}(\textrm{MAE}) = \frac{1}{M} \sum_{i=1}^M
    \underbrace{
    \left[
    \frac{1}{\sqrt{2\pi}\sigma_i} \int_\mathbb{R} |y_i - \hat y_i| e^{-\frac{(y_i - \overline y_i)^2}{2\sigma_i^2}} dy_i
    \right]}_{\displaystyle J}
\end{equation}
Let us consider for the calculation only $J$. The following change of variables can be used
\begin{equation}
    s = \frac{y_i-\overline{y}_i}{\sqrt{2}\sigma_i} \, \, \rightarrow ds = dy_i \frac{1}{\sqrt{2}\sigma_i}
\end{equation}
therefore
\begin{equation}
    \begin{array}{lll}
         J &  = & \displaystyle\int_\mathbb{R} |
            s\sqrt{2}\sigma_i - \underbrace{(\hat y_i-\overline{y}_i)}_{\displaystyle\delta_i}
         | e ^{-s^2} ds = \\
         &  = & \displaystyle \frac{1}{\sqrt{2}\pi} \displaystyle\int_\mathbb{R} |
         s\sqrt{2}\sigma_i - \delta_i| e^{-s^2}ds
    \end{array}
\end{equation}
due to the absolute value, the integral must be split into two parts: $J=J_A+J_B$. Part A for $s\sqrt{2}\sigma_i - \delta_i \geq 0$ and part B for $s\sqrt{2}\sigma_i - \delta_i < 0$. The two integrals are
\begin{equation}
    J_A = \displaystyle \frac{\sqrt{2}\sigma_i}{\sqrt{\pi}} \displaystyle\int_{(\delta_i / (\sqrt{2}\sigma_i))}^\infty \left(
         s - \frac{\delta_i}{\sqrt{2}\sigma_i}\right) e^{-s^2}ds
\end{equation}
and
\begin{equation}
    J_B = -\displaystyle \frac{\sqrt{2}\sigma_i}{\sqrt{\pi}} \displaystyle\int_\infty^{(\delta_i / (\sqrt{2}\sigma_i))} \left(
         s - \frac{\delta_i}{\sqrt{2}\sigma_i}\right) e^{-s^2}ds
\end{equation}
Let us start with $J_A$. To further simply the notation let us define
\begin{equation}
    \tilde \delta_i = \frac{\delta_i}{\sqrt{2}\sigma_i}
\end{equation}
so $J_A$ can now be evaluated
\begin{equation}
    \begin{array}{lll}
         J_A & = & \displaystyle \frac{\sigma_i \sqrt{2}}{\sqrt{\pi}} \left[
            \int_{\tilde \delta_i}^\infty s e^{-s^2}ds -
            \int_{\tilde \delta_i}^\infty \tilde \delta_i e^{-s^2}ds
         \right]\\[12pt]
         & = &\displaystyle \frac{\sigma_i \sqrt{2}}{\sqrt{\pi}} \left[
            \frac{1}{2} e^{-\tilde \delta_i^2} -
            \tilde \delta_i \frac{\sqrt{\pi}}{2} \textrm{erfc}(\tilde \delta_i) \right]\\[12pt]
        & = & \displaystyle \frac{\sigma_i }{\sqrt{2\pi}}
         e^{-\tilde \delta_i^2}-\frac{1}{\sqrt{2}} \delta_i \textrm{erfc}(\tilde \delta_i)
    \end{array}
\end{equation}
Analogously
\begin{equation}
    \begin{array}{lll}
         J_B & = & -\displaystyle \frac{\sigma_i \sqrt{2}}{\sqrt{\pi}} \left[
            \int_\infty^{\tilde \delta_i} s e^{-s^2}ds -
            \int_\infty^{\tilde \delta_i} \tilde \delta_i e^{-s^2}ds
         \right]\\[12pt]
         & = &\displaystyle \frac{\sigma_i \sqrt{2}}{\sqrt{\pi}} \left[
            \frac{1}{2} e^{-\tilde \delta_i^2} +
            \tilde \delta_i \frac{\sqrt{\pi}}{2} \textrm{erfc}(-\tilde \delta_i) \right]\\[12pt]
        & = & \displaystyle \frac{\sigma_i }{\sqrt{2\pi}}
         e^{-\tilde \delta_i^2}+\frac{1}{\sqrt{2}} \delta_i \textrm{erfc}(-\tilde \delta_i)
    \end{array}
\end{equation}
therefore, with some simplifications
\begin{equation}
\label{eq:J1}
    J=\frac{\sqrt{2}\sigma_i}{\sqrt{\pi}}\left[
      e^{-\tilde \delta_i^2}+ \sqrt{\pi} \tilde \delta_i
      \textrm{erf}(\tilde \delta_i)
      \right]
\end{equation}
Now this form is not easy to interpret, and it can be brought in a more interpretable form with some additional manipulation. Let us start by noticing that
\begin{equation}
\label{eq:erf1}
    \tilde \delta_i \textrm{erf}(\tilde \delta_i) = |\tilde \delta_i| \textrm{erf}(|\tilde \delta_i|)
\end{equation}
since $\tilde \delta_i \textrm{erf}(\tilde \delta_i) = -\tilde \delta_i \textrm{erf}(-\tilde \delta_i)$ due to the fact that $\textrm{erf}(-x)=-\textrm{erf}(x)$. Additionally, Equation (\ref{eq:erf1}) can be rewritten as
\begin{equation}
\label{eq:erf2}
    \tilde \delta_i \textrm{erf}(\tilde \delta_i) = |\tilde \delta_i| \textrm{erf}(|\tilde \delta_i|) = |\tilde \delta_i|
    (1-\textrm{erfc}(|\tilde \delta_i|))
\end{equation}
where $\textrm{erfc}(x)$ is the complementary error function. Using Equation (\ref{eq:erf2}), Equation (\ref{eq:J1}) can be rewritten as
\begin{equation}
    J=|\delta_i| +\frac{\sqrt{2}}{\sqrt{\pi}}\sigma_i e^{-\tilde \delta_i^2} - |\delta_i|\textrm{erfc}\left(
      \frac{|\delta_i|}{\sqrt{2}\sigma_i}
    \right)
\end{equation}
Now $\mathbb{E}(\textrm{MAE})$ can be finally written
\begin{equation}
\begin{array}{lll}
     \mathbb{E}(\textrm{MAE}) & = &  \displaystyle\frac{1}{M}\displaystyle\sum_{i=1}^M |\hat y_i-\overline{y}_i|+\frac{1}{M}\sum_{i=1}^M \left\{
      \frac{\sqrt{2}}{\sqrt{\pi}} \sigma_i e^{-\delta_i^2/(2\sigma_i^2)} +\right.\\
     & & -\left. |\delta_i| \textrm{erfc}\left(
        \displaystyle \frac{|\delta_i|}{\sqrt{2}\delta_i}
     \right)\right\}\\
\end{array}
\end{equation}
This concludes the derivation.

\section{Alternative Calculation of $\mathbb{E}(_ra)$}
\label{app:class}

The starting point of this alternative derivation is the formula
\begin{equation}
\label{eq:ea}
    \mathbb{E}(_ra) = \displaystyle \frac{1}{M} \sum_{i=1}^M \left[
        (1-q) \mathcal{A}(_ry_i)+ q \mathcal{A}(1-_r\hspace{-1mm}y_i) \right]
\end{equation}
where $\mathcal{A}(_ry_i)$ is
\begin{equation}
\label{eq:eq2}
    \mathcal{A}(_ry_i)=_r\hspace{-1mm}y_i H(\hat y_i-\alpha)+(1-_r\hspace{-1mm}y_i)(1- H(\hat y_i-\alpha))
\end{equation}
from Equation (\ref{eq:acc}). Equation (\ref{eq:ea}) can be expanded by using Equation (\ref{eq:eq2})  as
\begin{equation}
\begin{array}{lll}
    \mathbb{E}(_ra) & = &\displaystyle \frac{1}{M} \sum_{i=1}^M \left[
        (1-q) (_ry_i H(\hat y_i-\alpha)+(1-_r\hspace{-1mm}y_i)(1- H(\hat y_i-\alpha)) + \right. \\
     &   & \left. q ((1-_ry_i) H(\hat y_i-\alpha)+_r\hspace{-1mm}y_i(1- H(\hat y_i-\alpha)) \right] \\
     & = & \displaystyle \frac{1}{M} \sum_{i=1}^M \left[
        _ry_i H(\hat y_i-\alpha)-q\ _ry_i H(\hat y_i-\alpha) + (1-_r\hspace{-1mm}y_i)(1-H(\hat y_i-\alpha))+
        \right. \\
     & & \left. -q(1- _r\hspace{-1mm}y_i)(1-H(\hat y_i-\alpha)) + q(1-_r\hspace{-1mm}y_i)H(\hat y_i-\alpha)\right. \\
     & & \left.  +q\ _ry_i(1-H(\hat y_i-\alpha)) \right]
\end{array}
\end{equation}
After some algebra, this can be simplified and brought in the form
\begin{equation}
\label{eq:expect_acc2}
\begin{array}{lll}
     \mathbb{E}(_ra) & =& \displaystyle \frac{1}{M} \sum_{i=1}^M \left[
        (1-q)\underbrace{(1-_r\hspace{-1mm}y_i)(1-H(\hat y_i-\alpha))}_{\textrm{TN}}+(1-q)\underbrace{_ry_i H(\hat y_i-\alpha)}_{\textrm{TP}}+
     \right.\\
     & = & \left. q \underbrace{(1-_r\hspace{-1mm}y_i)H(\hat y_i-\alpha)}_{\textrm{FP}}+q \underbrace{_ry_i(1-H(\hat y_i-\alpha))}_{\textrm{FN}} \right]
\end{array}
\end{equation}
In Equation (\ref{eq:expect_acc2})  it is clearly indicated which part gives (when summed) the true positives (TP), true negatives (TN), false positives (FP), and false negatives (FN).
So Equation (\ref{eq:expect_acc2}) can be rewritten as
\begin{equation}
\label{eq:app1}
    \mathbb{E}(_ra) = (1-q) \frac{\textrm{TP}+\textrm{TN}}{M} + q \frac{\textrm{FP}+\textrm{FN}}{M}
\end{equation}
The final formula can be easily obtained by noting that
\begin{equation}
    \begin{array}{rll}
        a & = & \displaystyle \frac{\textrm{TP}+\textrm{TN}}{M} \\[0.5cm]
        1-a & = &  \displaystyle \frac{\textrm{FP}+\textrm{FN}}{M}
    \end{array}
\end{equation}
This concludes the derivation.

 \bibliographystyle{elsarticle-num}
 \bibliography{cas-refs}

\begin{thebibliography}{10}
\expandafter\ifx\csname url\endcsname\relax
  \def\url#1{\texttt{#1}}\fi
\expandafter\ifx\csname urlprefix\endcsname\relax\def\urlprefix{URL }\fi
\expandafter\ifx\csname href\endcsname\relax
  \def\href#1#2{#2} \def\path#1{#1}\fi

\bibitem{michelucci2019multi}
U.~Michelucci, F.~Venturini, Multi-task learning for multi-dimensional
  regression: application to luminescence sensing, Applied Sciences 9~(22)
  (2019) 4748.

\bibitem{wilkinson2022hybrid}
C.~J. Wilkinson, C.~Trivelpiece, R.~Hust, R.~S. Welch, S.~A. Feller, J.~C.
  Mauro, Hybrid machine learning/physics-based approach for predicting oxide
  glass-forming ability, Acta Materialia 222 (2022) 117432.

\bibitem{zhu2021machine}
Q.~Zhu, Z.~Liu, J.~Yan, Machine learning for metal additive manufacturing:
  predicting temperature and melt pool fluid dynamics using physics-informed
  neural networks, Computational Mechanics 67~(2) (2021) 619--635.

\bibitem{krishnan2018predicting}
N.~A. Krishnan, S.~Mangalathu, M.~M. Smedskjaer, A.~Tandia, H.~Burton,
  M.~Bauchy, Predicting the dissolution kinetics of silicate glasses using
  machine learning, Journal of Non-Crystalline Solids 487 (2018) 37--45.

\bibitem{carrasquilla2017machine}
J.~Carrasquilla, R.~G. Melko, Machine learning phases of matter, Nature Physics
  13~(5) (2017) 431--434.

\bibitem{morningstar2017deep}
A.~Morningstar, R.~G. Melko, Deep learning the ising model near criticality,
  Journal of Machine Learning Research (2018).

\bibitem{tanaka2017detection}
A.~Tanaka, A.~Tomiya, Detection of phase transition via convolutional neural
  networks, Journal of the Physical Society of Japan 86~(6) (2017) 063001.

\bibitem{baldi2016jet}
P.~Baldi, K.~Bauer, C.~Eng, P.~Sadowski, D.~Whiteson, Jet substructure
  classification in high-energy physics with deep neural networks, Physical
  Review D 93~(9) (2016) 094034.

\bibitem{de2016jet}
L.~de~Oliveira, M.~Kagan, L.~Mackey, B.~Nachman, A.~Schwartzman,
  Jet-images—deep learning edition, Journal of High Energy Physics 2016~(7)
  (2016) 1--32.

\bibitem{guest2016jet}
D.~Guest, J.~Collado, P.~Baldi, S.-C. Hsu, G.~Urban, D.~Whiteson, Jet flavor
  classification in high-energy physics with deep neural networks, Physical
  Review D 94~(11) (2016) 112002.

\bibitem{carrasco2013tpz}
M.~Carrasco~Kind, R.~J. Brunner, Tpz: photometric redshift pdfs and ancillary
  information by using prediction trees and random forests, Monthly Notices of
  the Royal Astronomical Society 432~(2) (2013) 1483--1501.

\bibitem{collister2007megaz}
A.~Collister, O.~Lahav, C.~Blake, R.~Cannon, S.~Croom, M.~Drinkwater, A.~Edge,
  D.~Eisenstein, J.~Loveday, R.~Nichol, et~al., Megaz-lrg: a photometric
  redshift catalogue of one million sdss luminous red galaxies, Monthly Notices
  of the Royal Astronomical Society 375~(1) (2007) 68--76.

\bibitem{ravanbakhsh2016estimating}
S.~Ravanbakhsh, J.~Oliva, S.~Fromenteau, L.~Price, S.~Ho, J.~Schneider,
  B.~P{\'o}czos, Estimating cosmological parameters from the dark matter
  distribution, in: International Conference on Machine Learning, PMLR, 2016,
  pp. 2407--2416.

\bibitem{carleo2019machine}
G.~Carleo, I.~Cirac, K.~Cranmer, L.~Daudet, M.~Schuld, N.~Tishby,
  L.~Vogt-Maranto, L.~Zdeborov{\'a}, Machine learning and the physical
  sciences, Reviews of Modern Physics 91~(4) (2019) 045002.

\bibitem{ghosh2022cautionary}
A.~Ghosh, B.~Nachman, A cautionary tale of decorrelating theory uncertainties,
  The European Physical Journal C 82~(1) (2022) 1--11.

\bibitem{pmlr-v37-menon15}
A.~Menon, B.~V. Rooyen, C.~S. Ong, B.~Williamson,
  \href{http://proceedings.mlr.press/v37/menon15.html}{Learning from corrupted
  binary labels via class-probability estimation}, in: F.~Bach, D.~Blei (Eds.),
  Proceedings of the 32nd International Conference on Machine Learning, Vol.~37
  of Proceedings of Machine Learning Research, PMLR, Lille, France, 2015, pp.
  125--134.
\newline\urlprefix\url{http://proceedings.mlr.press/v37/menon15.html}

\bibitem{cour2011learning}
T.~Cour, B.~Sapp, B.~Taskar, Learning from partial labels, The Journal of
  Machine Learning Research 12 (2011) 1501--1536.

\bibitem{pmlr-v119-zheng20c}
S.~Zheng, P.~Wu, A.~Goswami, M.~Goswami, D.~Metaxas, C.~Chen,
  \href{http://proceedings.mlr.press/v119/zheng20c.html}{Error-bounded
  correction of noisy labels}, in: H.~D. III, A.~Singh (Eds.), Proceedings of
  the 37th International Conference on Machine Learning, Vol. 119 of
  Proceedings of Machine Learning Research, PMLR, 2020, pp. 11447--11457.
\newline\urlprefix\url{http://proceedings.mlr.press/v119/zheng20c.html}

\bibitem{natarajan2013learning}
N.~Natarajan, I.~S. Dhillon, P.~K. Ravikumar, A.~Tewari, Learning with noisy
  labels, Advances in neural information processing systems 26 (2013)
  1196--1204.

\bibitem{pmlr-v119-yao20b}
Q.~Yao, H.~Yang, B.~Han, G.~Niu, J.~T.-Y. Kwok,
  \href{http://proceedings.mlr.press/v119/yao20b.html}{Searching to exploit
  memorization effect in learning with noisy labels}, in: H.~D. III, A.~Singh
  (Eds.), Proceedings of the 37th International Conference on Machine Learning,
  Vol. 119 of Proceedings of Machine Learning Research, PMLR, 2020, pp.
  10789--10798.
\newline\urlprefix\url{http://proceedings.mlr.press/v119/yao20b.html}

\bibitem{pmlr-v119-bahri20a}
D.~Bahri, H.~Jiang, M.~Gupta,
  \href{http://proceedings.mlr.press/v119/bahri20a.html}{Deep k-{NN} for noisy
  labels}, in: H.~D. III, A.~Singh (Eds.), Proceedings of the 37th
  International Conference on Machine Learning, Vol. 119 of Proceedings of
  Machine Learning Research, PMLR, 2020, pp. 540--550.
\newline\urlprefix\url{http://proceedings.mlr.press/v119/bahri20a.html}

\bibitem{7159100}
T.~Liu, D.~Tao, Classification with noisy labels by importance reweighting,
  IEEE Transactions on Pattern Analysis and Machine Intelligence 38~(3) (2016)
  447--461.
\newblock \href {https://doi.org/10.1109/TPAMI.2015.2456899}
  {\path{doi:10.1109/TPAMI.2015.2456899}}.

\bibitem{michelucci2021estimating}
U.~Michelucci, F.~Venturini, Estimating neural network’s performance with
  bootstrap: A tutorial, Machine Learning and Knowledge Extraction 3~(2) (2021)
  357--373.

\bibitem{taylor1997introduction}
J.~Taylor, Introduction to error analysis, the study of uncertainties in
  physical measurements, 1997.

\bibitem{michelucci2018applied}
U.~Michelucci, Applied Deep Learning, Springer, 2018.

\bibitem{hogg2010probability}
R.~V. Hogg, E.~A. Tanis, D.~L. Zimmerman, Probability and statistical
  inference, Pearson/Prentice Hall Upper Saddle River, NJ, USA:, 2010.

\end{thebibliography}





\end{document}